\documentclass[11pt,a4paper]{article}
\usepackage[hyperref]{acl2021}

\usepackage{times}
\usepackage{latexsym}

\usepackage[T1]{fontenc}

\usepackage[utf8]{inputenc}
\usepackage{graphicx}
\usepackage{subcaption}
\usepackage{multirow}
\usepackage{balance}

\usepackage{microtype}
\usepackage{booktabs}

\aclfinalcopy

\title{TellMeWhy: A Dataset for Answering Why-Questions in Narratives}

\author{Yash Kumar Lal \\
  Stony Brook University \\
  \texttt{ylal@cs.stonybrook.edu} \\\And
  Nathanael Chambers \\
  US Naval Academy \\
  \texttt{nchamber@usna.edu} 
  \AND
  Raymond Mooney \\
  University of Texas, Austin \\
  \texttt{mooney@cs.utexas.edu} \\\And
  Niranjan Balasubramanian \\
  Stony Brook University \\
  \texttt{niranjan@cs.stonybrook.edu}}

\begin{document}
\maketitle
\begin{abstract}
Answering questions about why characters perform certain actions is central to understanding and reasoning about narratives.
Despite recent progress in QA, it is not clear if existing models have the ability to answer ``why'' questions that may require commonsense knowledge external to the input narrative. 
In this work, we introduce \textbf{TellMeWhy}, a new crowd-sourced dataset that consists of more than 30k questions and free-form answers concerning why characters in short narratives perform the actions described.
For a third of this dataset, the answers are not present within the narrative. 
Given the limitations of automated evaluation for this task, we also present a systematized human evaluation interface for this dataset. 
Our evaluation of state-of-the-art models show that they are far below human performance on answering such questions.
They are especially worse on questions whose answers are external to the narrative, thus providing a challenge for future QA and narrative understanding research. 
\end{abstract}

\section{Introduction}

The actions people perform are steps of plans to achieve their desired goals.
When interpreting language, humans naturally understand the reasons behind described actions, even when the reasons are left unstated \cite{schank_script}. 
For NLP systems, answering questions about \emph{why} people perform actions in a narrative can test this ability. Answering such questions often requires filling the implicit gaps in the story itself.

Consider this narrative from ROCStories~\cite{mostafazadeh-etal-2016-caters}: 

\noindent
\begin{small}
Rudy was convinced that bottled waters all tasted the same. He went to the store and bought several popular brands. He went back home and set them all on a table. He spent several hours tasting them one by one. He came to the conclusion that they actually did taste different. 
\end{small}

Now try to answer the question, \emph{``Why did he go to the store and buy several popular brands?''}
The answer \emph{``he wanted to taste test''} is not explicit in the narrative and requires us to read between the lines to fill in the gaps \cite{norvig1987}. 
While humans can visualise and process the events in a story to hypothesize why they might have occurred \cite{Kintsch1978TowardAM}, current NLP systems fall well short of exhibiting similar capabilities. They are unable to adequately formulate the reasons behind actions in specific contexts. 

How can we get NLP models to reason about why actions are performed? One way is to consider theories like script learning \cite{schank_script, pichotta-mooney-2014-statistical} or learning from co-occurrence \cite{chambers-jurafsky-2009-unsupervised}. But they only partially capture this type of knowledge -- much like other forms of commonsense knowledge, the reasons for why actions are performed are often left implicit in text. Even though there are many large scale QA datasets, they rarely contain questions about {\it why} people perform actions.

Therefore, we introduce the TellMeWhy dataset, a collection of 30,519 such why-questions, each with 3 ``gold standard'' human answers.
Each record in TellMeWhy contains a short story, an associated question, and its 3 possible answers.

Further, we focus on enabling \emph{human} evaluation of this dataset; human evaluation is more reliable than automatic metrics to evaluate such systems \cite{celikyilmaz2020evaluation, gatt-krahmer}.
However, reliability of human judgment is substantially impacted by experimental setup \cite{novikova-etal-2018-rankme, santhanam-shaikh-2019-towards}.
There is little consensus on how human evaluations should be conducted, so results are often incomparable across evaluations.

To this end, we present a systematized evaluation framework on MTurk for the TellMeWhy text generation task -- and release the framework for future researchers.
The MTurk interface asks annotators to rate generated answers on their grammaticality and validity.
We show that with our interface human answers are judged to be of high quality (99\% grammatical, 96\% valid) with strong inter-annotator agreement at 0.88 Fleiss Kappa.
This indicates high agreement and also confirms the design of our interface.

Finally, we present baseline results for TellMeWhy and compare against our human ceiling.
We finetune two large language models that have proven to be effective for a variety of tasks, GPT-2 \cite{radford2019language} and T5 \cite{2020t5}, and a dedicated question answering model, UnifiedQA \cite{khashabi-etal-2020-unifiedqa}, to perform this task.
Human evaluation is performed on their outputs from independent test data.
All models significantly under-perform the human benchmark and are especially worse on questions where the answer cannot be simply copied over from text in the narrative. The results clearly demonstrate the difficulty for current models to convincingly answer such why-questions.

This paper's contributions are as follows: (1) we introduce TellMeWhy, a large dataset of English why-questions for narratives derived from ROCStories \cite{mostafazadeh-etal-2016-corpus} and CATERS \cite{mostafazadeh-etal-2016-caters} along with answers from 3 distinct humans, (2) a systematized human evaluation interface to calibrate model outputs consistently, and (3) show that current models are ill-equipped to perform this task.
We release the dataset and evaluation suite at \url{http://lunr.cs.stonybrook.edu/tellmewhy}.

\section{Related Work}

\subsection{Datasets containing why-questions}

Most of the datasets related to why-questions fall into one or more of the following categories: (1) very small size, (2) not focused on stories, or (3) focused on connecting known events instead of answering reasoning questions.

Some corpora of why-questions have been collected manually: corpora described in \citet{verberne-etal-2006-data} and \citet{verberne-etal-2007} both comprise fewer than 400 questions and corresponding answers (one or two per question) formulated by native speakers.
\citet{dunietz-etal-2020-test} demonstrate that it is important to define what we want models to comprehend when building datasets for machine reading comprehension (MRC) tasks.
They design templates of understanding corresponding to the four elements identified by \citet{zwaan-etal-1995-the}.
For 201 questions, they design multiple-choice questions derived from \cite{lai-etal-2017-race} to test understanding of different categories of events.
All of these are very small corpora that cannot be viably used to further a model's understanding of why-questions in stories.

\citet{higashinaka-isozaki-2008-corpus} extend an existing factoid QA system to answer why-questions by integrating corpus based features, calling it NAZEQA.
\citet{oh-etal-2012-question} extract a set of answer candidates from a web corpus, and perform re-ranking using SVMs to predict the right answer.
\citet{oh-etal-2019-open} use an adversarial learning framework to generate a vector representation from the passage to judge whether the passage actually answers the why-question.
These papers focus on Japanese news \cite{Fukumoto2007AnOO, oh-etal-2012-question}, including NTCIR-6, and most critically, all these datasets are very small.

Some prior work focuses on knowledge extraction, not the \emph{reasons} behind the actions.
\citet{mrozinski-etal-2008-collecting} built a corpus of why-questions related to Wikipedia articles. These were general knowledge questions with solicited answers from paid workers.
Dependency parsing can be used to rephrase why-questions into statements with a `because' prompt to elicit explanations from models \cite{nie-etal-2019-learning}.
PhotoshopQuiA \cite{dulceanu-etal-2018-photoshopquia} contains questions and answers specifically about Photoshop.

NarrativeQA \cite{kocisky-etal-2018-narrativeqa} provides a dataset of 1,567 stories (books and movie scripts) containing 46,765 wh-questions written and answered by human annotators.
Unfortunately, only 9.78\% are why-questions, which makes for a small collection.
QuAIL \cite{Rogers_Kovaleva_Downey_Rumshisky_2020} has a small subset of multiple choice questions pertaining to causality in user stories.
These datasets are targeted at broad abilities of reading comprehension, not specifically about explaining actions in stories.

\begin{table}[!t]
    \centering
    \begin{small}
    \begin{tabular}{|c|c|c|}
    \toprule
        Dataset & Size & Domain \\\midrule
        NTCIR-6 & 200 & Japanese news \\
        \citet{mrozinski-etal-2008-collecting} & 695 & Wikipedia \\
        PhotoshopQuiA & 2,854 & Product focused \\
        NarrativeQA & 4,573 & Books+Movie scripts \\
        \citet{dunietz-etal-2020-test} & 201 & Exam questions \\
    \bottomrule
    \end{tabular}
    \end{small}
    \caption{Previous why-question corpora. NarrativeQA has 46,765 questions of which 4,573 are why-questions.}
    \label{tab:rel_work}
\end{table}

Some recent datasets causally connect events in text, but they do not target answering why-questions.
ATOMIC \cite{Sap_Le_Bras_Allaway_Bhagavatula_Lourie_Rashkin_Roof_Smith_Choi_2019} consists of entries that describe a likely cause/effect of events.
Most notably, ATOMIC is \emph{non-contextual} so it is more about general knowledge, not interpreting a specific story/context.
Perhaps most relevant is GLUCOSE \cite{mostafazadeh-etal-2020-glucose}, a crowdsourced dataset of implicit commonsense knowledge in the form of causal mini-theories grounded in narrative context.
These theories are semi-structured inference rules.
This dataset is not aimed at answering why-questions, but at creating direct relationships between events already mentioned in the story.
They focus on capturing specific cause-enable type relations.
Annotators were given a very constrained task -- they had to select options from a drop down menu describing inference rules.

Abductive commonsense reasoning tests whether models can come up with a plausible explanation to connect a set of events.
\citet{Bhagavatula2020Abductive} present ART with two abductive tasks: 1) given two observations, select one out of two plausible hypotheses, 2) and generate text connecting two events.
This line of work focuses on connecting the dots between two events and does not address explaining {\it why} an action was performed. Our work crucially differs from these because the answer is often not in the story at all.
StrategyQA \cite{geva-etal-2021-strategyqa} is a new dataset focusing on performing better implicit reasoning for multi-hop question answering tasks.

We summarize the different why-questions corpora in \autoref{tab:rel_work}.
None of them represent a large dataset focused on answering why-questions about actions in a narrative.

\subsection{Human evaluation for NLG tasks}

Among language generation tasks, machine translation has received the most attention in terms of human evaluation.
Qualified crowd workers score output translations given the source or reference text to calibrate MT systems \cite{sakaguchi-van-durme-2018-efficient, graham-etal-2013-continuous, graham-etal-2014-machine}.
WMT conducts annual evaluation of outputs of systems submitted to the shared task and uses it as one of the primary metrics (along with BLEU) to rank systems \cite{bojar-etal-2016-findings, bojar-etal-2017-findings, bojar-etal-2018-findings, barrault-etal-2019-findings, barrault-etal-2020-findings}.

ChatEval \cite{sedoc-etal-2019-chateval} is an evaluation platform for chatbots.
\citet{zellers2020evaluating} present a leaderboard for their advice generation task.
These platforms incorporate some manual analysis, but focus on very different tasks.
None of their Mechanical Turk interfaces can be used for our task.
We were unable to find a consistent interface for human evaluation of an open-ended question answering task.
To address this flaw, we propose a standard human intelligence task (HIT) evaluation scheme for our dataset.

\section{Dataset Creation}
\label{sec:data_create}

We want to test the abilities of models to understand the reasoning behind actions in a story. 
Therefore, we create a dataset of \emph{why} questions that ask for explanations for actions performed in a story. 
Answering these questions requires an understanding of the events that are explicit in the story as well as access to implicit common-sense knowledge on how people use actions as parts of plans to achieve goals. 
To cover a wide-range of common situations, we utilize ROCStories \cite{mostafazadeh-etal-2016-corpus}, a collection of 45,496 five-sentence commonsense stories. We also develop a small ``hidden'' test set that was only used for the final evaluation using the CATERS \cite{mostafazadeh-etal-2016-caters} subset of ROCStories.

\subsection{Why-Question Generation}
\label{subsec:generate}
Our strategy for creating {\it why} questions is simple. For each action in the narrative, we formulate a why question by applying simple template-based transformations. We dependency parse each sentence using SpaCy's en\_core\_web\_sm model \cite{spacy}.
We use the generated parse tree to rephrase the sentence into a question about the action described.
The generated parse tree is used to extract the subject, object, and verb.
We consider 3 types of sentences and design question templates accordingly: (1) sentences that have a primary and auxiliary verb, (2) sentences that only have a primary verb, and (3) sentences that only contain an auxiliary verb.
For the first, the question template is: ``Why \{aux\_verb\} \{subject\} \{verb\} \{object\}?", for the second: ``Why did \{subject\} \{verb\_lemma\} \{obj\}?", and for the third; ``Why \{aux\_verb\} \{subject\} \{obj\}?".

This procedure yielded a little over 113k questions from ROCStories, and 489 questions from the CATERS portion. We selected at random 32,165 questions from stories that had at least three questions\footnote{Since we ask annotators to read an entire story to answer these questions, avoiding stories with fewer questions optimizes reading time.}. We ensure that there is no overlap between the two subsets.

\begin{table}[!t]
    \centering
    \begin{tabular}{|c|c|c|}
    \toprule
        Split & \# stories & \# questions \\
        \midrule
        Train & 7558 & 23964 \\
        Dev & 944 & 2992 \\
        Test & 944 & 3099 \\
        \midrule
        Hidden Test & 190 & 464 \\
        \midrule
        Total & 9,636 & 30,519 \\
        \bottomrule
    \end{tabular}
    \caption{Dataset Statistics}
    \label{tab:dataset_stats}
\end{table}

\subsection{Collecting Answers}
\label{subsec:curate}
\begin{figure*}
\centering
\begin{subfigure}[t]{\columnwidth}
  \includegraphics[width=\columnwidth]{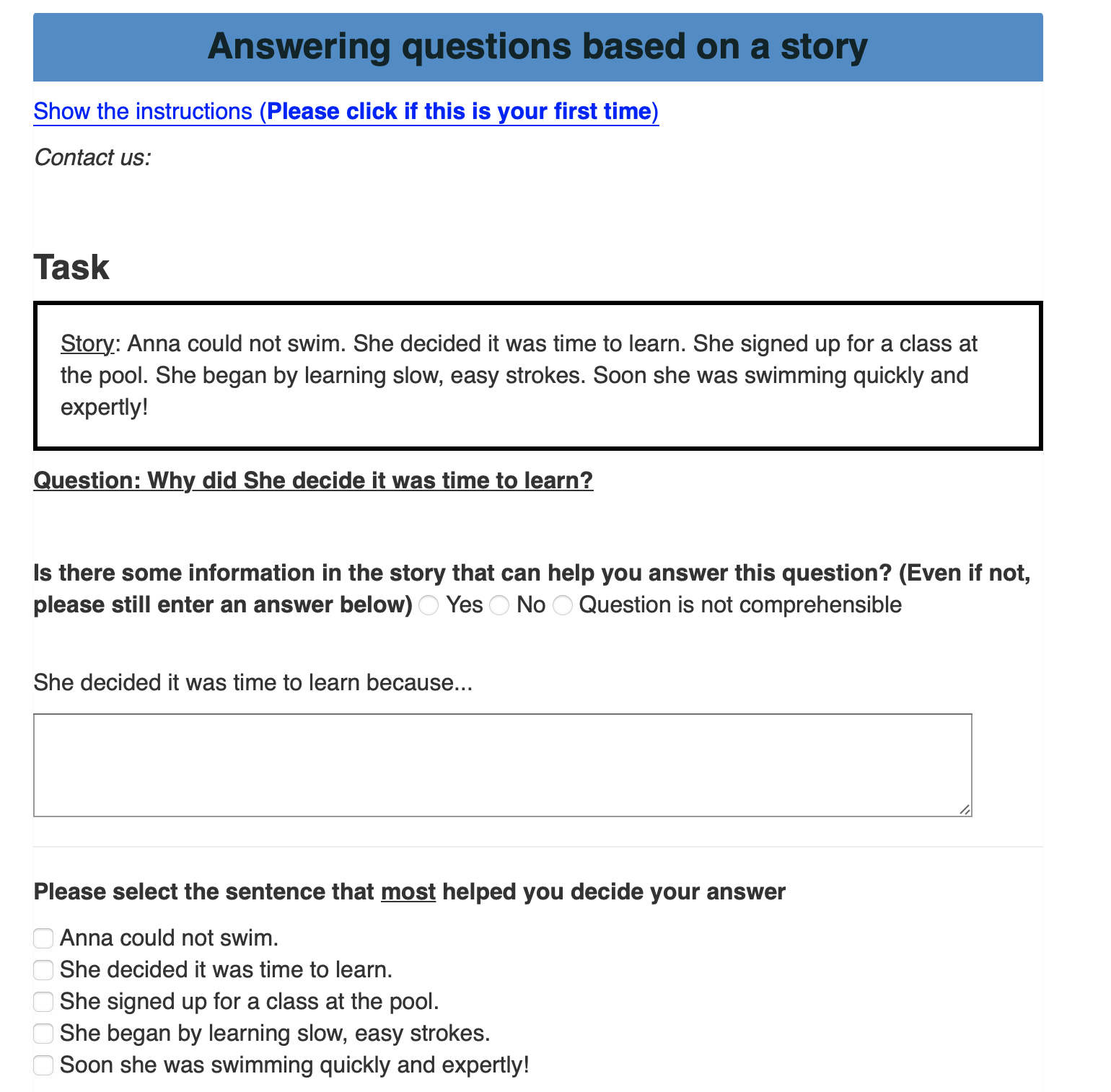}
  \caption{Task 1: Answer collection}
  \label{fig:ans_collect} 
\end{subfigure}
\begin{subfigure}[t]{\columnwidth}
  \includegraphics[width=\columnwidth]{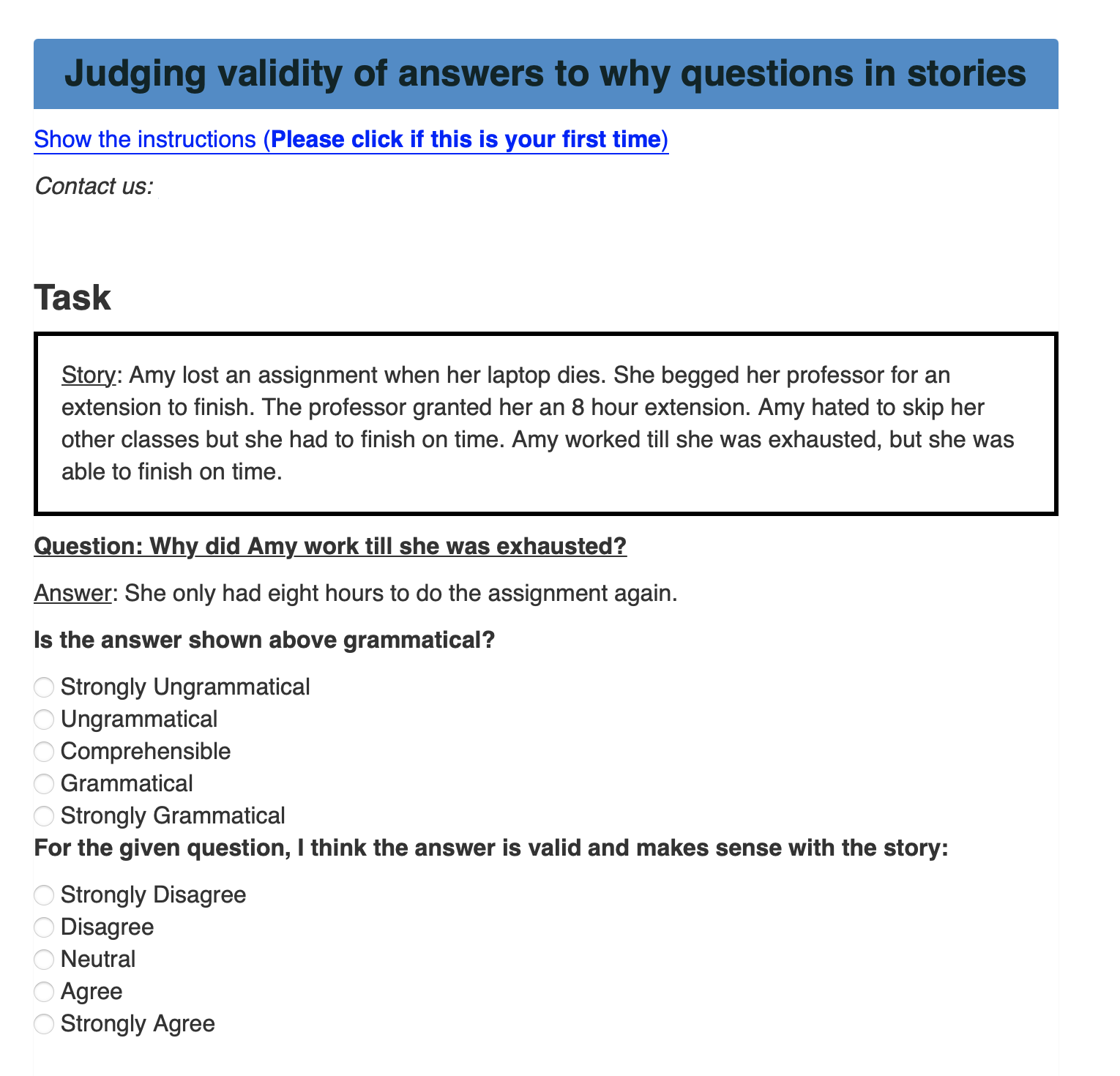}
  \caption{Task 2: Answer validation}
  \label{fig:crowdsourcing-interface}
\end{subfigure}
\caption{MTurk interfaces used to curate data from crowd-source workers}
\end{figure*}

We crowd-sourced answers to these questions using Amazon Mechanical Turk. 
Figure~\ref{fig:ans_collect} shows the interface used to collect these answers.
Annotators were presented a narrative and asked to answer three {\it why} questions in free-form. 
For each question, they were also asked to provide judgments about the comprehensibility of the question, and whether the narrative explicitly contained the answer.
They also selected the sentences from the narrative which influenced their answer (if any). 
To avoid variability in answer prefixes, we provide a prompt to start answering the question. We rephrase the sentence from which the question was generated to create these prompts.
We consider the same categorisation of sentences described in \autoref{subsec:generate}.
For sentences that have both primary and auxiliary verbs, the answer prompt is of the form: ``\{subject\} \{aux\_verb\} \{verb\} \{object\} because...".
When it only contains a primary verb, it is of the form: ``\{subject\} \{verb\} \{object\} because...".
If it only contains an auxiliary verb, it is of the form: ``\{subject\} \{aux\_verb\} \{object\} because...".
We found, over several iterations of this HIT, that providing a prompt gave workers an initial direction and improved the quality of answers collected.

We ask three distinct annotators (three-way redundant task) to answer each of these questions.
Annotators are not allowed to copy pieces of text to make up an answer.
We discard questions that were deemed incomprehensible by any annotator.\footnote{On ROCStories we discarded 1,546 questions and on CATERS we discarded 25 questions}
With this process, we obtained 3 answers each from 30,055 questions from 9,636 stories (see \autoref{sec:data} for more details).
\autoref{tab:dataset_stats} shows basic statistics of the dataset.
We refer to annotations from the CATERS data as the hidden test set.
Examples of records in the dataset are presented in \autoref{tab:data_ex}.
The narrative does not explicitly contain an answer for the second question.
We call these types the implicit-answer questions; they require extra common-sense inference to produce a plausible answer.
Questions are categorised as implicit-answer if at least 2 out of 3 human annotators indicate that the answer cannot be explicitly found in the narrative.
The annotators indicated as much and, based on their commonsense knowledge, provided plausible answers.

\begin{table}[!t]
    \centering
    \begin{tabular}{|p{0.97\linewidth}|}
    \toprule
        \textbf{Story:} Sandra got a job at the zoo. She loved coming to work and seeing all of the animals. Sandra went to look at the polar bears during her lunch break. She watched them eat fish and jump in and out of the water. She took pictures and shared them with her friends. \\
        \textbf{Question:} Why did Sandra go to look at the polar bears during her lunch break? \\
        \textbf{Ans:} she wanted to take some pictures of them. \\
        \midrule
        \textbf{Story:} Cam ordered a pizza and took it home. He opened the box to take out a slice. Cam discovered that the store did not cut the pizza for him. He looked for his pizza cutter but did not find it. He had to use his chef knife to cut a slice. \\
        \textbf{Question:} Why did Cam order a pizza? \\
        \textbf{Ans:} Cam was hungry. \\
    \bottomrule
    \end{tabular}
    \caption{TellMeWhy examples. The first is answerable directly from text in the story, but the second requires external knowledge. We only show one out of three available answers here.}
    \label{tab:data_ex}
\end{table}

\subsection{Validating Answers}
\label{subsec:validate}

To ensure an even higher-quality test set, we conducted another round of crowdsourcing to validate the answers by the first set of crowd-workers on the CATERS portion (464 questions). This validation interface is show in Figure~\ref{fig:crowdsourcing-interface}. It also serves as the base design for our systematized human evaluation.
Annotators are presented a story, a related question, and the three answers that were collected as described in Section \ref{subsec:curate}. 

Three new annotators then rated two aspects of each answer: \\
\noindent(1) \textbf{Grammaticality} -- Workers are asked to rate the grammaticality of each answer on 5-point Likert scales, ranging from `Strongly Ungrammatical' to `Strongly Grammatical'.
An answer is strongly grammatical if it follows all the rules of English grammar.
It is grammatical if there is a mistake in tense, number, punctuation or something minor.
It is comprehensible if there are clear grammatical mistakes but its meaning can be inferred, and it is then 
considered to be neutral on the Likert scale.

\noindent(2) \textbf{Validity} -- Workers are asked to rate the validity of each answer on a 5-point Likert scale.
Given the story and question, the annotators check if the given answer `is valid and makes sense with the story'.
An answer is considered invalid if it does not give a plausible reason relevant to the question asked and instead states irrelevant information.

Annotators agreed (by majority) that 99.07\% of answers are grammatical and 95.47\% of answers are valid.
On grammaticality, there is some disagreement in judgment 0.7\% of the time, while there is some disagreement in judgment 1\% of the time for answer validity.
We measured the inter-rater reliability of annotators' judgments using weighted Fleiss’s Kappa \cite{Marasini2016AssessingTI} and follow the weighting scheme used by \citet{bastan-etal-2020-authors}.
This measure has a penalty for each dissimilar classification based on the distance between two classes.
For instance, if two annotators classify a document as a positive, the agreement weight is 1, but if one classifies as a positive, and the other classifies as slightly positive the agreement weight is less.
The weighted agreement score for this subset is 0.88 for grammaticality annotations and 0.81 for validity annotations, indicating that the annotations are highly reliable.
More details can be found in Appendix \ref{subsec:fleiss}.

\section{Dataset Analysis}

One of the key distinguishing aspects of answering {\it why} questions is that, in addition to understanding explicitly stated events, they also require access to commonsense explanations that may be external to the narrative. We conduct some analyses to investigate the prevalence of this phenomenon:
(i) We asked annotators to judge whether the answer to a question could be found stated explicitly or only implicitly in the narrative and find that at least two out of three annotators could not find explicit answers in the story 28.82\% of the time. (ii) We also asked crowd-workers to indicate which sentences helped them answer the question. Out of 91,557 collected answers, we find that 39,661 answers were provided without an influential sentence from the story.
(iii) Last, we observe that there is only a 57.04\% lexical overlap between the words used in answers and the original narrative. This suggests that annotators included new inferred information in their answers, instead of just copying something from the story. We calculate lexical overlap as the number of common tokens in the narrative and the answer divided by the length of the answer.

We hypothesize that questions about the first action in a story  are more difficult to answer since there is no prior information to provide an explicit answer.
We find that 55.03\% of such questions were judged to be implicit-answer questions
by a majority of the assigned annotators.
Such questions help test systems' ability to infer plausible answers rather than just copy answers from the text.

\begin{table*}[!t]
    \centering
    \begin{tabular}{|c|c|c|c|c|c|}
    \toprule
        Evaluated on & Model & BLEU & RG-L F1 & BLEURT & BertScore \\
        \midrule
        \multirow{5}{*}{Full Test Set} & GPT-2-OO & 4.45 & 0.07 & -1.23 & -0.55 \\
        & GPT2-FT & 3.96 & 0.13 & -0.75 & 0.18 \\
        & T5-OO & 9.89 & 0.13 & -0.963 & 0.23 \\
        & T5-FT & \textbf{24.53} & 0.24 & \textbf{-0.28} & \textbf{0.48} \\
        & UnifiedQA & 21.97 & \textbf{0.25} & -0.30 & 0.43 \\
        \midrule
        \multirow{5}{*}{\shortstack{Implicit-Answer Qs \\ in Test Set}} & GPT-2-OO & 4.45 & 0.06 & -1.22 & -0.54 \\
        & GPT2-FT & 3.89 & 0.12 & -0.805 & 0.17 \\
        & T5-OO & 8.14 & 0.11 & -0.99 & 0.22 \\
        & T5-FT & \textbf{16.31} & 0.17 & -0.51 & \textbf{0.38} \\
        & UnifiedQA & 14.6 & \textbf{0.18} & \textbf{-0.50} & 0.34 \\
        \bottomrule
    \end{tabular}
    \caption{Performance of models on the full test set and on implicit-answer questions in the test set using automated metrics. RG-L denotes ROUGE-L. The OO suffix denotes the vanilla version of the model while the FT version denotes the finetuned version.}
    \label{tab:auto_metrics}
\end{table*}

We also evaluated the diversity of the answers for each question using simple lexical overlap.
Of the 30k questions, only 150 questions had over 90\% overlap in all 3 answers, i.e., essentially, the 3 distinct annotators wrote the same answer.
For 4,243 other questions, two out of three answers had over 90\% overlap.
But for the vast majority of 26,068 questions, we obtained 3 fairly diverse answers.
The average overlap between them is 26.12\%.
On average, the answers were 7.59 words long.

Overall, this analysis indicates how TellMeWhy differs from prior datasets. The answers cannot always be retrieved or connected to other events in the given text. 

\section{Benchmarking}
How well do large language models answer {\em why} questions on narratives and what are their failure modes?  To answer these, we use TellMeWhy to benchmark the performance of multiple state-of-the-art models and provide an analysis of their performance.

Formally, given a story \textit{S} as context and a related why-question \textit{Q}, models are required to generate a plausible answer \textit{A} for the question. Since the answers are open-ended texts we compare them on standard automatic evaluation metrics for generation but also conduct a human evaluation.  

\subsection{QA Models}

\noindent \textbf{GPT-2} \cite{radford2019language} is a large transformer-based language model trained on an enormous web corpus, which has been shown to be effective on a wide-range of language related tasks including question answering.
It was one of the first models trained on diverse data to outperform domain-specific language models.

We used Huggingface \cite{wolf-etal-2020-transformers} to finetune a pretrained GPT-2 model on TellMeWhy.
As input, the model receives a concatenation of the narrative and the related question (in that order), and the target is the answer.
The input and target are separated using the `[SEP]' token.
We finetune the model with batch size 16, learning rate 1e-5 and maximum output length 25.
The model is trained until the dev loss fails to improve for 3 iterations.

\noindent \textbf{T5} \cite{2020t5} is an encoder-decoder model pre-trained on a mixture of unsupervised and supervised tasks in a multi-task setting, where each task is converted into a text-to-text format.
It is a text-to-text model, which means it can be trained on arbitrary tasks involving textual input and output.
T5 has achieved the state of the art on many natural language understanding (NLU) tasks.
More details about hyperparameter sweeps can be found in \autoref{sec:sweep}.

We finetuned a pretrained T5-base model from HuggingFace \cite{wolf-etal-2020-transformers} on TellMeWhy.
Since it is a natural language generation task related to a story, we use the SQuAD format specified in Appendix D.15 of \citet{2020t5} to format our inputs.
Our narrative serves as the `context' and the why-question is used as the `question' in the selected input format.
We train the model with batch size 16, learning rate 5e-5, maximum source length 75 and maximum answer length 30.
The model is trained until the dev loss fails to improve for 3 iterations.

\noindent \textbf{UnifiedQA} \cite{khashabi-etal-2020-unifiedqa} is a single pre-trained model that performs well across 20 different question answering datasets.
It is built on top of a T5 model and  simplifies finetuning by unifying the various formats used by T5.
Its ability to perform both extractive and abstractive QA tasks makes it a suitable candidate for calibrating this task.
A pretrained version of this model is available via HuggingFace \cite{wolf-etal-2020-transformers} under the name ``allenai/unifiedqa-t5-base".
The input format for this model is simple, just requiring the question and the narrative to be separated by a newline symbol.
We train this model using learning rate 1e-5 (same as the original paper) and retain other hyperparameters from finetuning T5 as described above.

\subsection{Automatic Evaluation}

We evaluate all of the above models on both the test set and the hidden test set (questions from CATERS data).
For automatic evaluation, we report BLEU \cite{papineni-etal-2002-bleu}, ROUGE-L \cite{lin-2004-rouge}, BLEURT \cite{sellam-etal-2020-bleurt} scores using the bluert-base-128 checkpoint, and BertScore \cite{Zhang2020BERTScore:} using the default roberta-large checkpoint.
These numbers are presented in \autoref{tab:auto_metrics}.

We select one human answer at a time and (using SacreBLEU \cite{post-2018-call}) calculate the BLEU scores for model output with all three references, and select the maximum.
Since BLEURT is a sentence level metric, to calculate the reported BLEURT, we average all the (output, reference) scores to obtain a corpus score for each reference.
We then select the maximum BLEURT corpus score over all 3 human references.
It is important to note that BLEURT was proposed as a metric for relative comparison, not absolute calibration.
We also report BertScore F1\footnote{idf and rescale\_with\_baseline flags are set to True. idf is calculated over all gold answers in the test set.} \cite{Zhang2020BERTScore:} as another semantic automatic evaluation metric.
We report a max BertScore in the same way as BLEURT and BLEU: by taking the maximum score of the model output with each human answer taken one at a time.

Vanilla model results are obtained by loading an existing pretrained model from HuggingFace and running inference with the input formats described above. They are not trained on TellMeWhy. We see that vanilla pretrained models are unable to perform this task at all.
Finetuning a pretrained model results in improvements since it better models the relationship between the story, the question, and a possible answer.
On the full test set, the finetuned T5 model performs the best on our task.
In \autoref{tab:auto_metrics}, we also see that models perform a lot worse on implicit-answer questions. 
\subsection{Human Evaluation}

For open-ended text generation tasks like answering why-questions, the absence of an automatic evaluation that correlates well with human judgments is a major challenge \cite{chen-etal-2019-evaluating, ma-etal-2019-results, caglayan-etal-2020-curious,howcroft-etal-2020-twenty}.

We conduct a human evaluation on the hidden test set with a standardized interface to compare different models. 
We want to measure whether a model produces coherent and grammatical output and more importantly, whether the produced output is a valid answer for the given question. 
Our validation HIT~\autoref{subsec:validate} showed a way to conduct human evaluation of answers provided by other crowd-workers. 
We modified this HIT design to evaluate generated answers from models. 
For a given question, we present just one answer from a single model and then ask the crowd-workers to assess its grammaticality and validity.

For each story, question, and a model's answer, we ask 3 distinct annotators to provide judgments about grammaticality and validity.
This serves as the human evaluation interface for our task.
A sample HIT can be seen in \autoref{fig:crowdsourcing-interface}.

\begin{figure}[!t]
    \centering
    \includegraphics[width=\columnwidth]{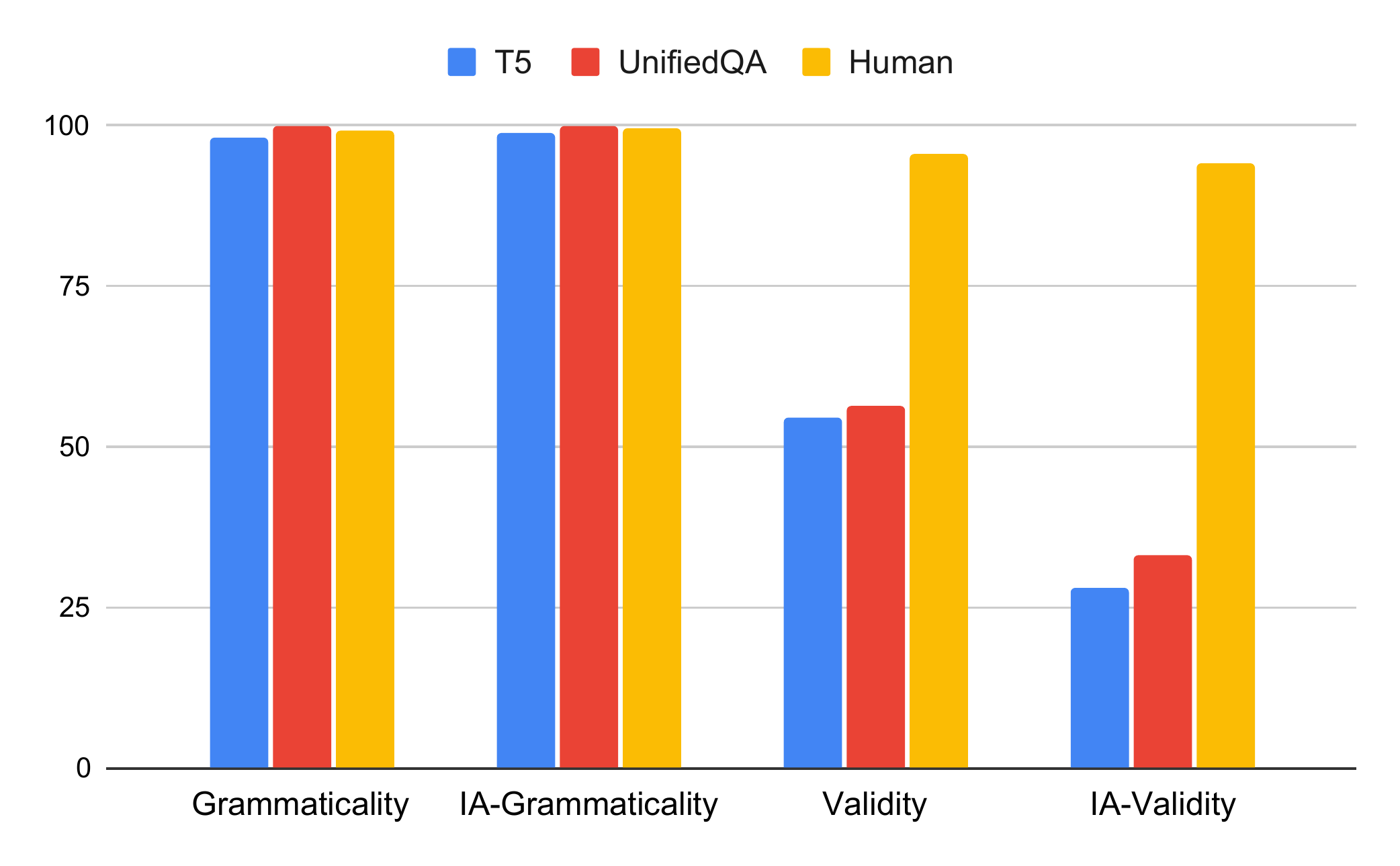}
    \caption{Human evaluated performance of answers. The IA prefix denotes performance on implicit-answer questions in the data.}
    \label{fig:human_eval_perf}
\end{figure}

We perform human evaluation of the fine-tuned versions of T5 and UnifiedQA, the two models that performed the best on automatic metrics.
We evaluate the outputs of these models on the hidden test set.
We calculated inter-annotator agreement for these judgments using the method described in \autoref{subsec:validate}, and they were >80\%, indicating high agreement.
The models mostly produce grammatical answers, but fail to adequately explain many actions in the story.
\autoref{fig:human_eval_perf} shows that, under human evaluation, models significantly under-perform humans at producing valid answers to why-questions.
Models fare worse when the answers to the questions are external to the narrative.

\begin{figure}[t!]
    \centering
    \includegraphics[width=\columnwidth]{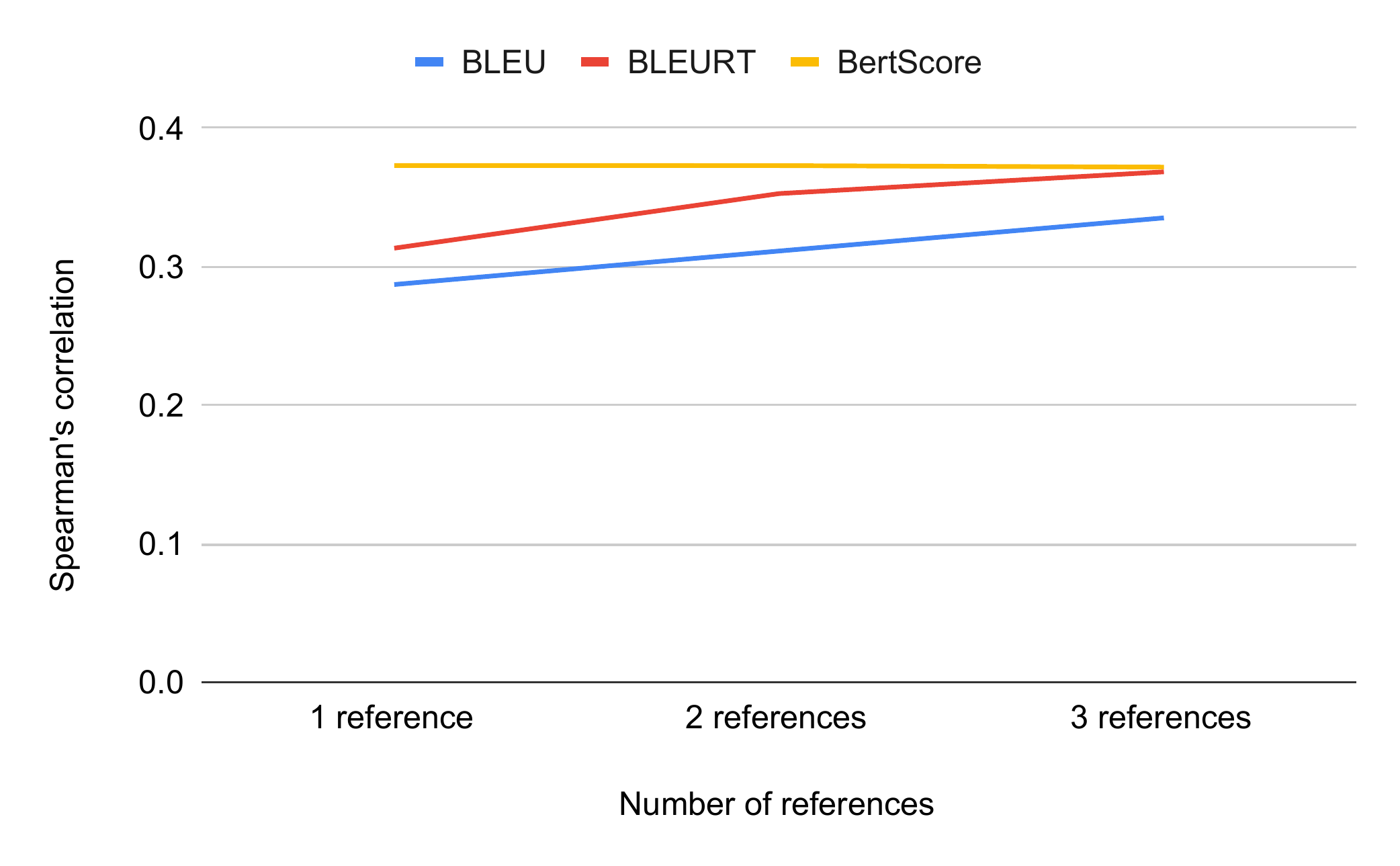}
    \caption{Correlation of automatic metrics with human validity judgment of model outputs. For each question, we have 3 crowdsourced human answers available to us. We selected a number of human answers randomly (X-axis) and calculated scores for each model output across different automatic metrics. Finally, we obtained Spearman's correlation (Y-axis) of these scores in comparison with Likert judgments provided by annotators for each human answer.}
    \label{fig:val_corr}
\end{figure}

Human evaluation is slow and expensive, so we performed a correlation analysis between the automated metrics and human judgments to gauge usefulness of popular automated metrics.
\autoref{fig:val_corr} shows that the embedding-based metrics are only weakly correlated with human validity judgments, while lexical metrics did even worse.
None of the automatic metrics show a strong correlation, confirming our earlier assertion that human evaluation is the most appropriate way to analyze model performance on this open-ended generation task.
BertScore and BLEURT have weak correlation with human validity judgments.
All three metrics improve their correlation with human judgments slightly as the number of human reference answers is increased; however, the increase is not large.

Our human judgment interface can serve as a standard human evaluation of any future model's performance on our dataset, and we will make code available for automatically generating HITs for evaluating the outputs of any model.
This standardized evaluation approach is similar in spirit to GENIE \cite{khashabi2021genie}, a contemporary work that also presents an evaluation framework for a large set of generation tasks.

\begin{figure}[t!]
\centering
\vspace{-1.5em} 
\begin{subfigure}{\columnwidth}
  \includegraphics[width=\columnwidth]{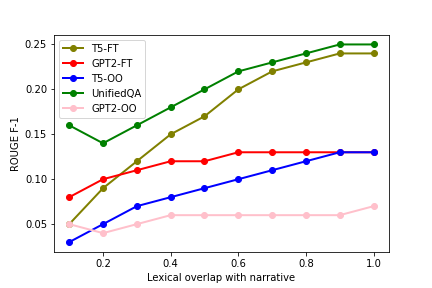}
  \caption{ROUGE F-1 trend}
  \label{fig:rouge-f-trend} 
\end{subfigure}
\begin{subfigure}{\columnwidth}
  \includegraphics[width=\columnwidth]{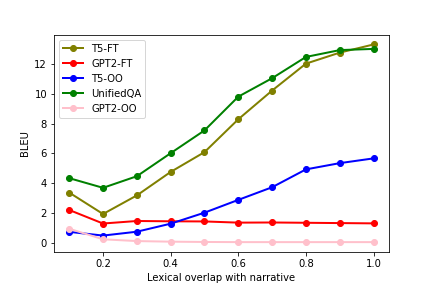}
  \caption{BLEU trend}
  \label{fig:bleu-trend}
\end{subfigure}
\caption{Model performance on different metrics with change in lexical overlap between a question's answers (as provided by humans) and the related narrative.}
\end{figure}

\subsection{Analysis}

In order to better understand when models are generating valid answers, we analyzed the correlation between model performance and a proxy for checking when human provided answers were in the input narrative. To this end, we aligned ROUGE F-1 scores
with the lexical overlap of human answers and the
story text. \autoref{fig:rouge-f-trend} shows how ROUGE F-1 scores for our models increases as the lexical overlap also increases between the answers and corresponding story.
The same is presented for BLEU in \autoref{fig:bleu-trend}.
Perhaps not surprisingly, this empirically shows that models do best when the answer is in the text, and suffer greatly when it is not (implicit answers).
This further illustrates the value of TellMeWhy, as well as its challenge, that standard models are largely incapable of performing the reasoning needed to produce plausible answers that are assumed common knowledge by the story writer.

\autoref{tab:model_copy} also shows that the best performing models mainly learnt to copy complete or parts from the narrative to generate answers, treating this largely as an extractive task. On average, more than three-fourths of T5 and UnifiedQA's answers are based on words in the narrative text. T5 is worse compared to UnifiedQA in terms of copying, with a much larger fraction of questions (59.44\% vs 27.44) with high lexical overlap (i.e.\ lexical overlap > 90\%). In comparison, the average narrative overlap for human answers is much lower than the best-performing models, since people are able to infer answers that are not in the text. If the models are to successfully answer {\em why} questions, they need to look beyond copying texts.

\begin{table}[!t]
    \centering
    \begin{tabular}{|c|c|c|}
    \toprule
        System & Copied ans & Avg overlap \\
        \midrule
        Vanilla GPT2 & 0\% & 23.09\% \\
        Finetuned GPT2 & 0\% & 28.94\% \\
        Vanilla T5 & 5.50\% & 53.71\% \\
        Finetuned T5 & 59.44\% & 85.91\% \\
        UnifiedQA & 27.44\% & 76.51\% \\
        \midrule
        Human Answers & 35.03\% & 57.04\% \\
    \bottomrule
    \end{tabular}
    \caption{Overlap between answers and the original narrative. This indicate how much original text models produce.}
    \label{tab:model_copy}
\end{table}

\section{Conclusion}

This paper introduces a large, novel QA dataset, \textbf{TellMeWhy}, containing questions about {\it why} characters in a narrative perform their depicted actions.
This challenge problem complements the variety of existing QA datasets, addressing the scarcity of ``why'' questions.
Using both automated metrics and human evaluation, we show that existing deep-learned language models perform quite poorly at answering such questions. We also illustrate the uniqueness of this challenge where the answer is sometimes in the story itself, but often not, thus requiring a richer model that can draw on commonsense knowledge or external reasoning abilities.

We believe that progress on answering such questions requires new systems that can reason about actions, plans, and goals in order to achieve a deeper understanding of narrative text, as was initially argued over four decades ago  \cite{Schank1977ScriptsPG}. We hope that TellMeWhy encourages further research in this area.

\section*{Acknowledgement}

This material is based on research that is supported in part by the Air Force Research Laboratory (AFRL), DARPA, for the KAIROS program under agreement number FA8750-19-2-1003 and in part by the National Science Foundation under the award IIS \#2007290.
The authors would like to thank the anonymous reviewers and the area chair for their feedback on this work.
We would also like to thank Horace Liu for helping us run some experiments for the camera ready version.

\bibliography{anthology,custom}
\bibliographystyle{acl_natbib}

\clearpage

\appendix

\section{Hyperparameter Sweep}
\label{sec:sweep}

We describe the hyperparameters and the range of values we experimented with.
The best hyperparameters are chosen on the basis of model loss on the validation set.
For both GPT-2 \cite{radford2019language} and T5 \cite{2020t5}, we conduct guided sweeps for learning rate, batch size and epochs.
We experiment with 1e-5, 5e-5 and 1e-4 for learning rate.
Batch sizes of 8, 16 and 32 were tried.
Models were trained for 20, 30 and 50 epochs, and we found that models converged between 30 and 50 epochs.
In the case of T5, we also experiment with different lengths of inputs and target outputs.
We trained models with maximum source lengths of 50, 60 and 75 tokens.
For target length, we experimented with 15, 25 and 30 tokens.
The maximum output length is treated as a hyperparameter for GPT-2, and we tried 15, 20, 25 and 30 tokens.

\section{Dataset Creation}
\label{sec:data}

The method described in \autoref{subsec:generate} creates 489 questions from the 200 stories in the CATERS dataset -- 36 stories with 1 question, 63 with 2, 59 with 3, 30 with 4, and 6 with 5. We collect 3 human answers for all questions.
For ROCStories, this creates 113,213 questions from 45,496 stories -- 7,555 stories with 1 question, 13,431 with 2, 13,349 with 3, 7356 with 4, and 1865 with 5.
We randomly select 32,165 questions from stories with 3 or 5 questions, for ease and efficiency of collecting annotations.
This is the smallest number for which we could gather 3 answers for at least 30,000 questions, which is a reasonable-sized dataset for training or fine-tuning large NLP models.

\section{Mechanical Turk tasks}

\subsection{Instructions}
\label{subsec:mturk_inst}

\begin{figure*}
\centering
\begin{subfigure}{\columnwidth}
  \includegraphics[width=\columnwidth]{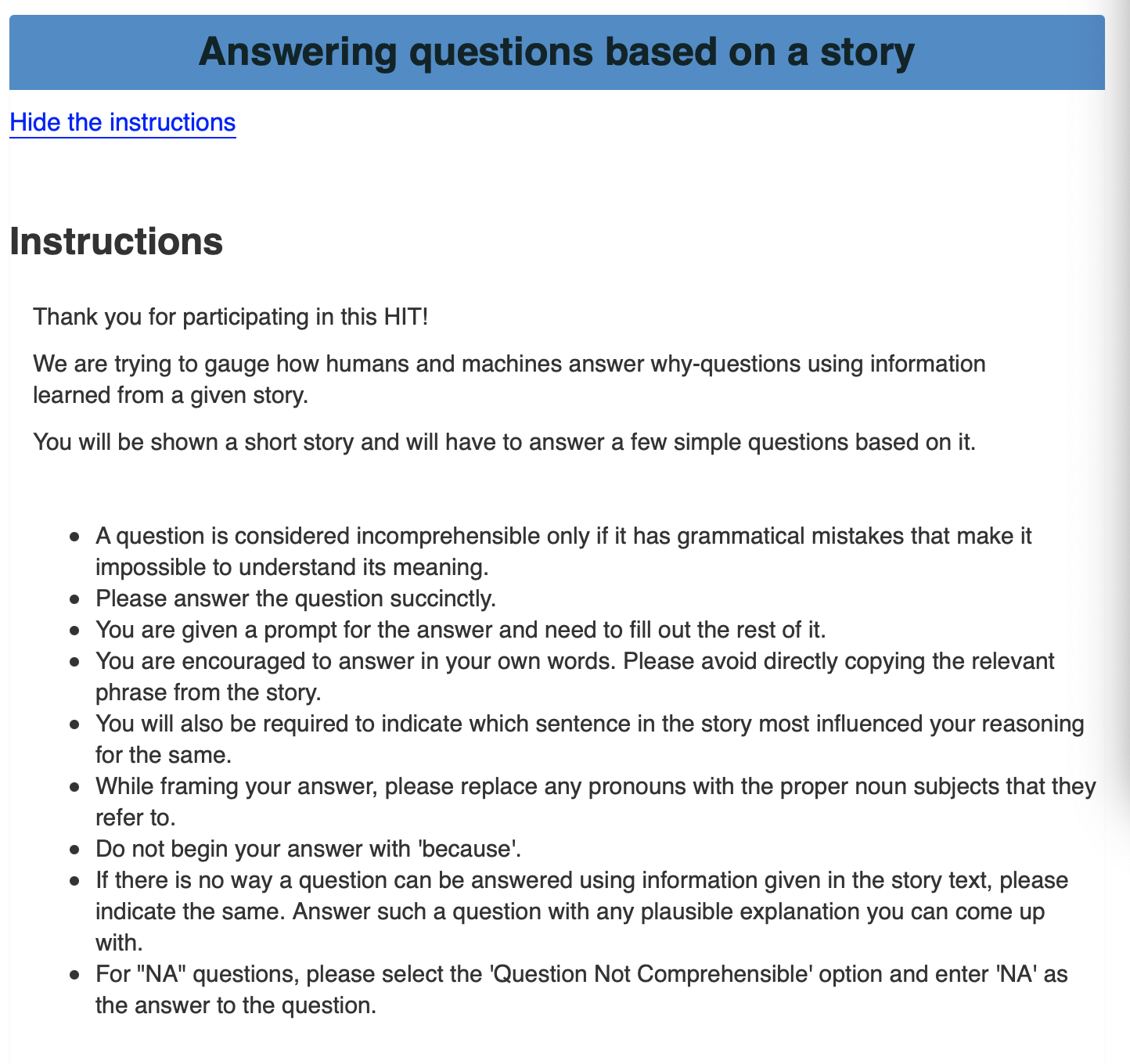}
  \caption{Instructions for answer collection task}
  \label{fig:ans_collect_inst} 
\end{subfigure}
\begin{subfigure}{\columnwidth}
  \includegraphics[width=\columnwidth]{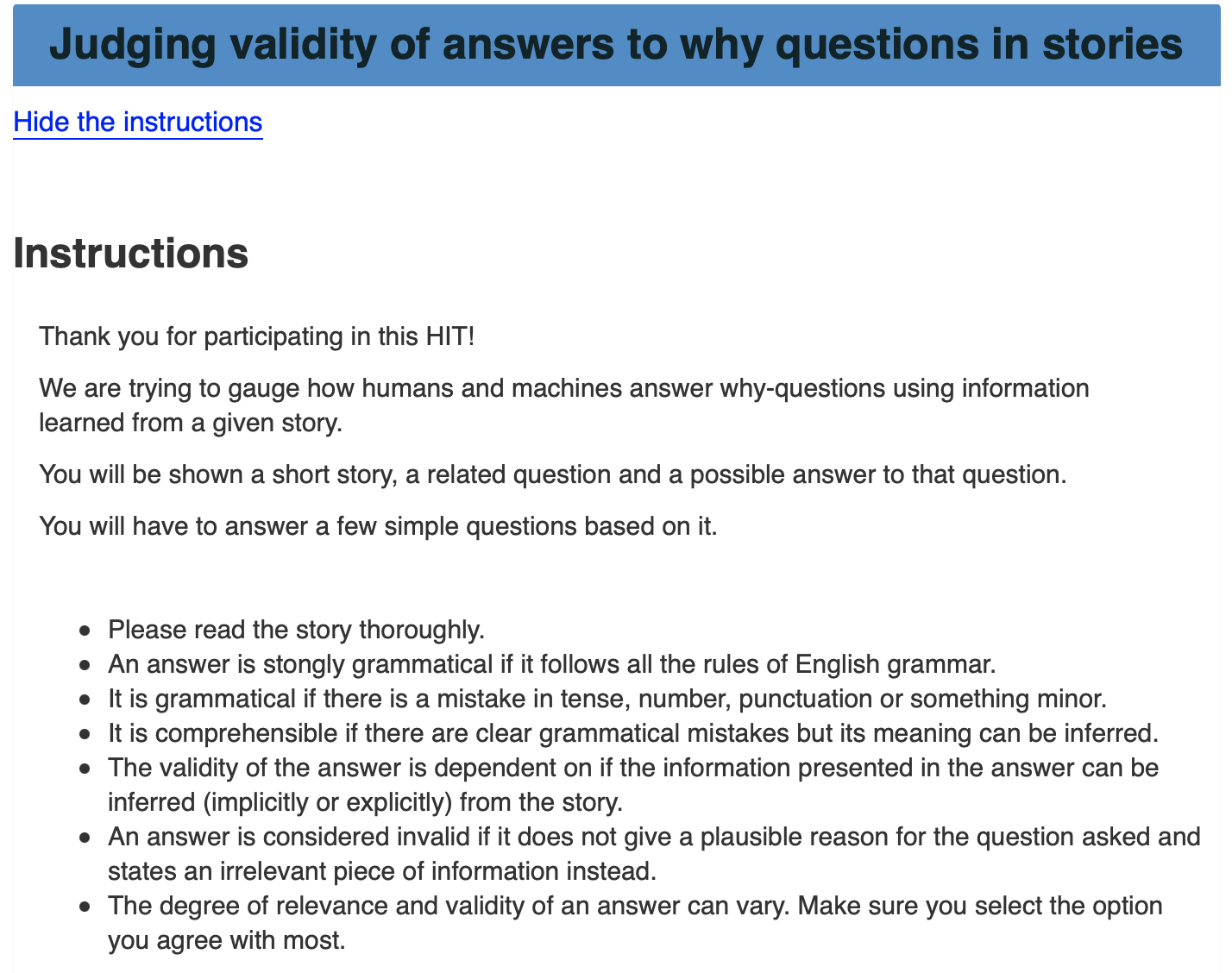}
  \caption{Instructions for answer validation task}
  \label{fig:ans-val-inst}
\end{subfigure}
\caption{Instructions for MTurk tasks}
\label{fig:inst}
\end{figure*}

We present the instructions given to annotators for both the tasks in \autoref{fig:inst}.
Annotators were given clear direction for both tasks.
We restricted both tasks to master turkers.
The second task (answer validity) was also used a sanity check for answers collected in the first task (answer collection).
Using results of the answer validity task (mentioned in \autoref{subsec:validate}), we see that humans provided high quality answers in the answer curation task.

\subsection{Inter-annotator agreement}
\label{subsec:fleiss}

We use weighted Fleiss Kappa to calculate inter-rater reliability.
The weights between different classes are shown in \autoref{tab:fleiss_kappa_weights} where negative, slightly negative, neutral, slightly positive, and positive classes are shown with -2, -1, 0, 1, and 2.
We follow the setup used in \citet{bastan-etal-2020-authors} for a similar multi-class labeling task.

\begin{table}[!tbh]
    \scriptsize
    \centering
    \begin{tabular}{|c|c|c|c|c|c|}
    \toprule
         & -2 & -1 & 0 & 1 & 2 \\
        \midrule
        -2 & 1 & cos $\pi$/8 & cos $\pi$/4 & cos 3$\pi$/8 & 0 \\
        -1 & cos $\pi$/8 & 1 & cos $\pi$/8 & cos $\pi$/4 & cos 3$\pi$/8 \\
        0 & cos $\pi$/4 & cos $\pi$/8 & 1 & cos $\pi$/8 & cos $\pi$/4 \\
        1 & cos 3$\pi$/8 & cos $\pi$/4 & cos $\pi$/8 & 1 & cos $\pi$/8 \\
        2 & 0 & cos 3$\pi$/8 & cos $\pi$/4 & cos $\pi$/8 & 1 \\
    \bottomrule
    \end{tabular}
    \caption{Inter class weights used for computing inter annotated agreement}
    \label{tab:fleiss_kappa_weights}
\end{table}

\end{document}